\documentstyle[spie]{article} 
\input{psfig}   

\title{\vspace*{-0.5cm} \hspace*{\fill}{\normalsize LA-UR-00-2839}
\\[1.5ex] 
Geometric Morphology of Granular Materials} 

\author{Bernd R. Schlei\supit{a}, Lakshman Prasad\supit{b} and Alexei N. Skourikhine\supit{b} 
\skiplinehalf 
\supit{a}Los Alamos National Laboratory, Theoretical Division, T-1, MS E541,\\
Los Alamos, NM 87545, U.S.A.
\\
\supit{b}Los Alamos National Laboratory, Nonproliferation and 
International Security, NIS-7, MS E541,\\ Los Alamos, NM 87545, U.S.A.
}

\authorinfo{Further author information: (Send correspondence to B.R.S.)\\
B.R.S.: E-mail: schlei@lanl.gov\\ 
L.P.: E-mail: prasad@lanl.gov\\ 
A.N.S.: E-mail: alexei@lanl.gov}

\begin{document} 
  \maketitle 

\begin{abstract}
We present a new method to transform the spectral pixel information of 
a micrograph into an affine geometric description, which allows us to 
analyze the morphology of granular materials. We use spectral and 
pulse-coupled neural network based segmentation techniques to generate blobs, 
and a newly developed algorithm to extract dilated contours. A constrained 
Delaunay tesselation of the contour points results in a triangular mesh. 
This mesh is the basic ingredient of the Chodal Axis Transform, 
which provides a morphological decomposition of shapes. Such decomposition 
allows for grain separation and the efficient computation of the 
statistical features of granular materials.
\end{abstract}



\section{INTRODUCTION}

Mathematical morphology was born in 1964, when G. Matheron \cite{Matheron65} 
was asked to investigate the relationships between the geometry of porous 
media and their permeabilities, and when at the same time J. Serra
\cite{Serra65} was asked to quantify the petrography of iron ores, in order 
to predict their milling properties. This initial period (1964 -- 1968) has 
resulted in a first body of theoretical notions. The notion of a geometrical 
structure, or texture, is not purely objective. It does not exist in the 
phenomenon itself, nor in the observer, but somewhere between the two.
Mathematical morphology quantifies this intuition by introducing the
concept of structuring elements. Chosen by the morphologist, they interact
with the object under study, modifying its shape and reducing it to a sort 
of caricature which is more expressive than the actual initial phenomenon.
The power of this approach, but also its difficulty, lies in this structural
analysis.

In this paper, we are going to present a new method of structural analysis
for images (micrographs) of granular materials to the immediate benefit of
the field of micromechanics of materials. In this particular field of 
physics, hydrodynamic model calculations are performed to investigate the 
propagation of shockwaves through such heterogeneous media. To be specific, 
one needs to setup the calculations with very accurate initial conditions, 
e.g., distributions of grain sizes in case one is interested in the mechanical
properties of granular materials. In our presentation we demonstrate how to 
prepare the micrographs for extraction of proper input statistics data by 
using low and high level image processing algorithms. 

We perform image segmentation of a micrograph (original image and its smoothed
version) by applying either pulse coupled neural networks (PCNN)
\cite{SkourikhineThis} or a spectral segmentation. Contour extraction of the 
segmented blobs will be performed with a newly developed algorithm, which is 
highly parallel and therefore fast. The contours and its points form the input 
to the Chordal Axis Transform \cite{Prasad97,PrasadThis} (CAT), 
which provides a morphological decomposition of grain shapes into simplicial 
chain complexes of limbs and torsos. This decomposition is used to perform 
geometric filtering operations such as grain separation based on grain 
morphology. Subsequent to such filtering, we explain how the metrical and 
statistical properties of the granular material can be efficiently computed to 
obtain input for the hydrodynamic model calculations.

\newpage

   \begin{figure}[t]
   \begin{center}
   \vspace*{-0.5cm}
   \begin{tabular}{c}
   \psfig{figure=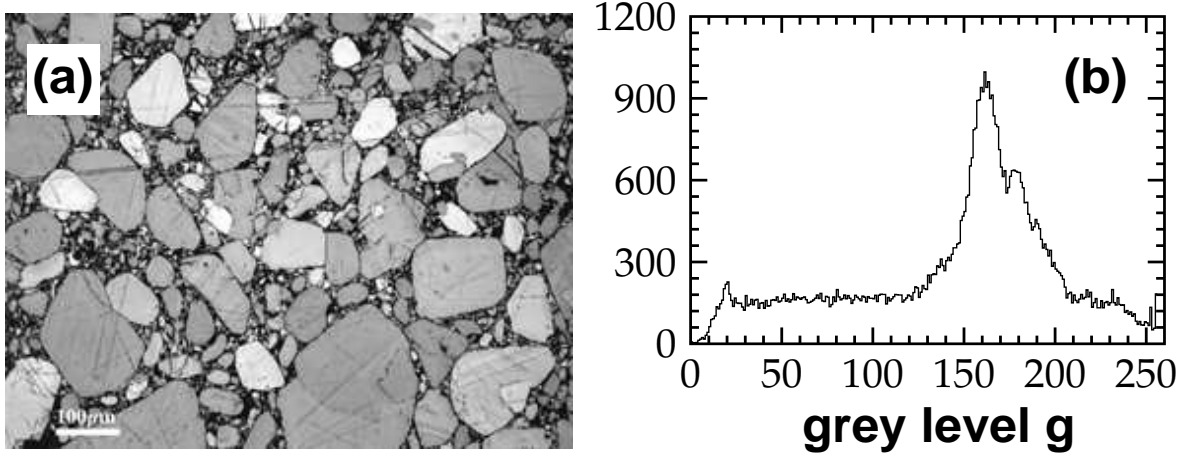,width=16cm} 
   \end{tabular}
   \vspace*{-0.5cm}
   \end{center}
   \caption[example] { 	  
   (a) original image,
   (b) corresponding pixel grey-level histogram. 
   } 
   \end{figure} 

\section{IMAGE PROCESSING}

In this paper, we discuss the image analysis of a micrograph (cf. Fig. 1.a) 
of an explosive material, PBX 9501 (95\% HMX and 5\% polymeric binder,
by weight). More images similar to the one shown in Fig. 1.a can be found in 
the paper by Skidmore \cite{Skidmore97} et al.  

\subsection{Image Segmentation and Image Smoothing} 

An image can in general be viewed as the spacial arrangement (positions)
of pixels, which contain spectral information (such as grey-levels, etc.).
In Fig 1.b we show the grey-level histogram of the explosive material, which
is shown in Fig 1.a. We can identify the pixels, which have rather
low grey-levels (dark) in the range of 0--100, with the polymeric binder,
and the remaining pixels (light) with the HMX crystals (grains). 
If we assign to the binder pixels black pixels and to the grain pixels white 
pixels, respectively, then this spectral segmentation leads to the bi-level 
image shown in Fig. 3.a.

A more sophisticated way of performing image segmentation can be acchieved
by applying a PCNN \cite{SkourikhineThis}. PCNN's use, simultaneously, the
spacial and spectral information content of an image for its segmentation
into blobs. We have applied such a PCNN to Fig 1.a. The result is shown
in Fig. 3.b (for more detail, cf. the paper by Skourikhine 
\cite{SkourikhineThis} et al.). In addition to segmentation, PCNN's can also
help to smooth an image, in order to suppress noise generated by the imager
or to remove undesired texture. We have smoothed Fig. 1.a while using a PCNN.
The resulting image 
 
   \begin{figure}[b]
   \begin{center}
   \vspace*{-0.5cm}
   \begin{tabular}{c}
   \psfig{figure=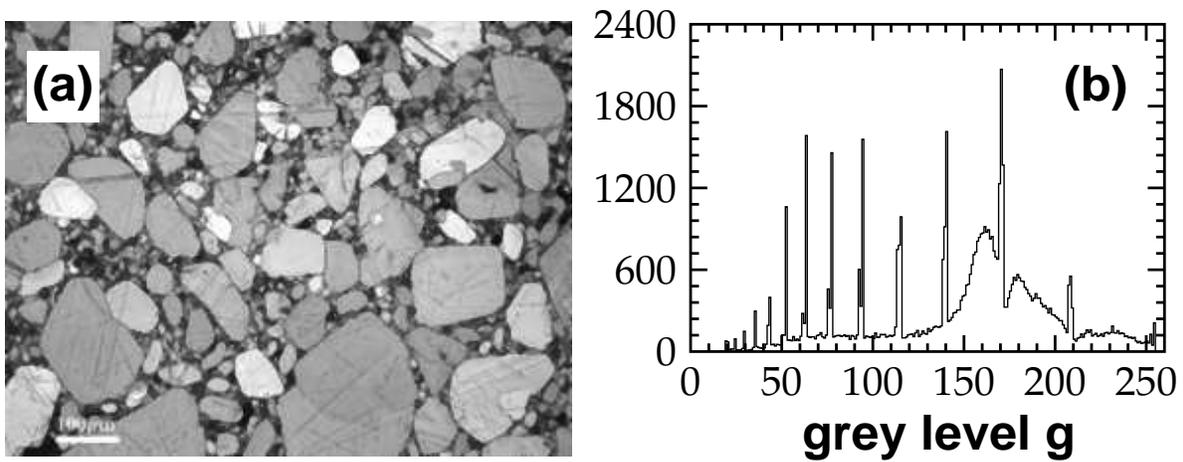,width=16cm} 
   \end{tabular}
   \vspace*{-0.5cm}
   \end{center}
   \caption[example] { 	  
   (a) smoothed image,
   (b) corresponding pixel grey-level histogram. 
   } 
   \end{figure} 

   \begin{figure}[t]
   \begin{center}
   \vspace*{-0.5cm}
   \begin{tabular}{c}
   \psfig{figure=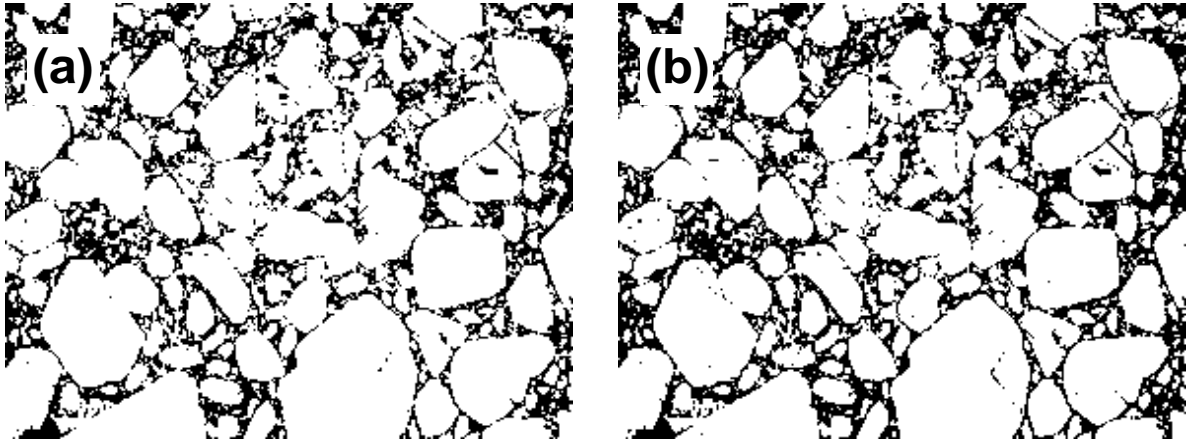,width=16cm} 
   \end{tabular}
   \vspace*{-0.5cm}
   \end{center}
   \caption[example] { 	  
   original image segmented by using
   (a) spectral range,
   (b) PCNN, (see text). 
   } 
   \end{figure} 
 
and its corresponding pixel grey-level histogram are
displayed in Fig. 2. We would like to emphasize, that the application of
a PCNN leads to smoothing without blurring the image \cite{SkourikhineThis}.
In Fig. 4. we have applied spectral and PCNN segmentation, respectively, as 
described above to the smoothed image Fig. 2.a. In total we have generated
four different segmented versions of the original image, where apparantly the 
level of pixel noise decreases when going from Fig. 3.b over Fig. 3.a and 
Fig. 4.a to Fig. 4.b. In the following, we shall use the four segmented 
images for their further morphological analysis. This section also 
demonstrates, that image analysis is not purely objective, as we indicated 
in the introduction. 

\subsection{Contour Extraction} 

The next step after the image segmentation is the contour extraction of
the blobs. For the material science application under consideration we desire 
blob contours, which are nondegenerate, i.e., they always enclose an area
larger than zero, and they never cross or overlap each other. Furthermore,
the contours should be oriented with respect to the objects and their possible
holes.  

In Fig. 5.a we show for the sake of illustration a cropped subimage of
Fig. 1.a with increased magnification. In Fig. 5.b we have applied a
spectral segmentation of the pixels grey-levels as described in the previous
section. Finally, Fig. 5.c shows the dilated contours as generated by the
algorithm \cite{Diconex1}, which is described in the paper by Schlei et al. 
\cite{Schlei00} and which fulfills all of the above stated requirements for 
the contours. 

   \begin{figure}[b]
   \begin{center}
   \vspace*{-0.5cm}
   \begin{tabular}{c}
   \psfig{figure=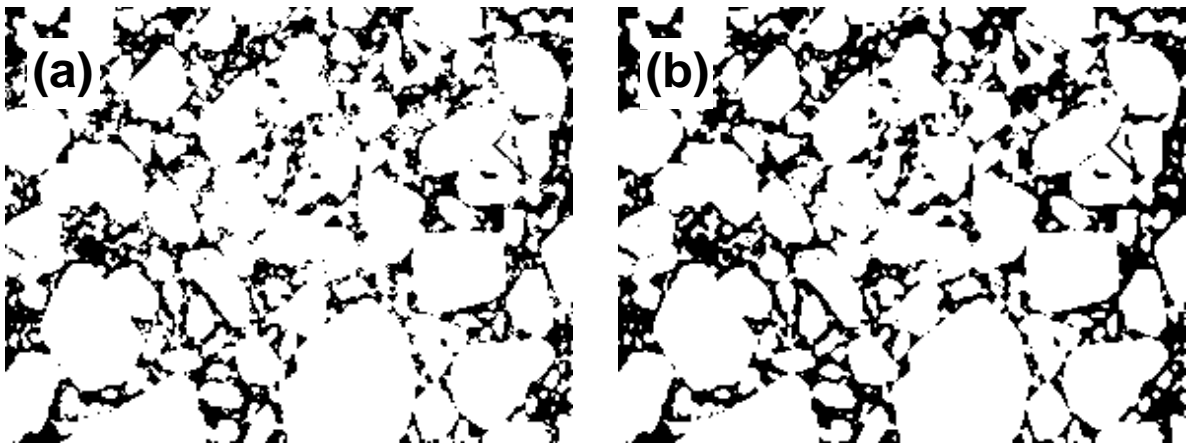,width=16cm} 
   \end{tabular}
   \vspace*{-0.5cm}
   \end{center}
   \caption[example] { 	  
   smoothed image segmented by using
   (a) spectral range,
   (b) PCNN, (see text). 
   } 
   \end{figure} 
 
   \begin{figure}[t]
   \begin{center}
   \vspace*{-0.2cm}
   \begin{tabular}{c}
   \psfig{figure=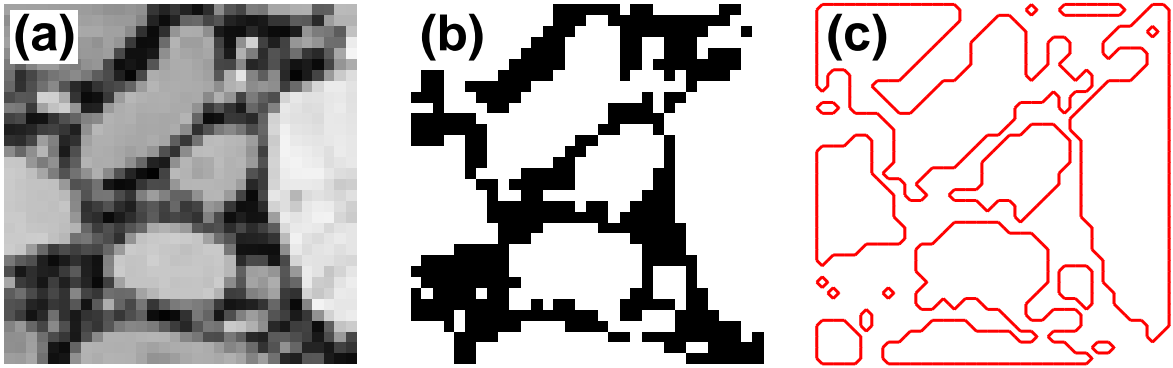,width=16.2cm} 
   \end{tabular}
   \vspace*{-0.5cm}
   \end{center}
   \caption[example] { 	  
   (a) cropped original image,
   (b) spectral segmentation,
   (c) dilated contours, (see text). 
   } 
   \end{figure} 
 
\subsection{Constrained Delaunay Tesselation and Chordal Axis 
Transform} 

The Constrained Delaunay Tesselation (CDT) of a simple planar polygon
(contour) is a decomposition of a polygon into triangles, such that the 
circumcircle of each triangle contains no vertex of the polygon inside
it that is simultaneously visible to two vertices of the triangle 
\cite{Prasad97}. Note, that the CDT of the contours is the key step that 
allows for the shape feature extraction. In Fig. 6.a we transform through
the CDT the spectral pixel information of our given micrograph into an affine 
geometric description. As a next step we use the pointset of the dilated
oriented contours and the generated triangular mesh with its neighbor
information of the triangles to compute the CAT skeleton 
\cite{Prasad97,PrasadThis} of the grains (cf. Fig. 6.b). Each arc of the
skeleton represents a simplical chain complex of either a limb or a torso,
respectively. A limb is a chain complex of pairwise adjacent triangles,
which begins with a junction triangle and ends with a termination triangle. 
A torso is a chain complex of pairwise adjacent triangles, which begins and 
ends with a junction triangle\cite{Prasad97,PrasadThis}.

\subsection{Grain Separation} 

In the micrograph (cf. Fig. 1.a) many grains might be very close to each 
other spatially such that they are connected through a single CAT skeleton. 
This is not satisfactory, if one wants to use the CAT skeleton directly 
for the computation of material statistics, because the number of grains is
unequal to the number of skeletons. If we compute for each CDT triangle
the average grey level of the pixels which are covered by it, we obtain
Fig. 6.b. For each given torso of the CAT we can investigate the fluctuation
of grey levels for its simplical chain complex. If two grains are connected
in the CAT skeleton through a torso, we would expect a larger grey level
fluctuation compared to the case, where a torso would be located within a 
single grain. In the following, we have used a threshold of 30
 
   \begin{figure}[b]
   \begin{center}
   \vspace*{-0.2cm}
   \begin{tabular}{c}
   \psfig{figure=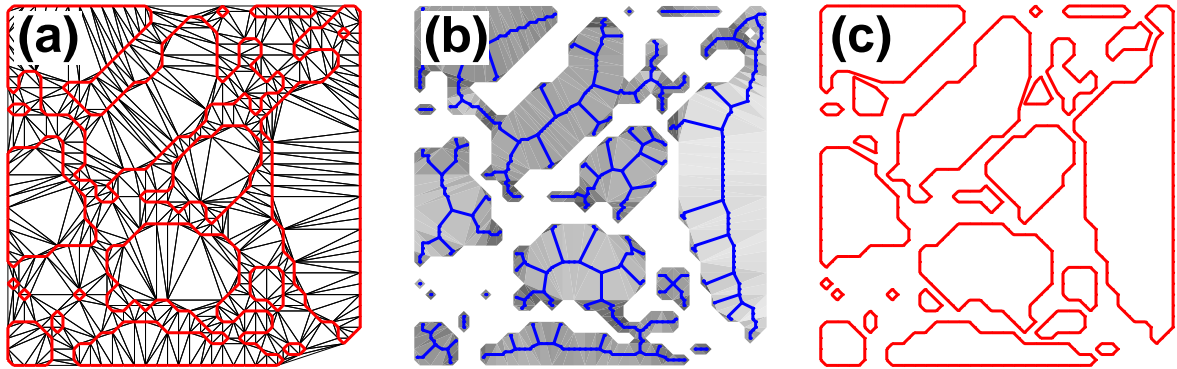,width=16.2cm} 
   \end{tabular}
   \vspace*{-0.5cm}
   \end{center}
   \caption[example] { 	  
   (a) Delaunay tesselation,
   (b) grey triangles and unpruned CAT skeleton,
   (c) refined contours, (see text). 
   } 
   \end{figure} 
 
   \begin{figure}[t]
   \begin{center}
   \vspace*{-0.2cm}
   \begin{tabular}{c}
   \psfig{figure=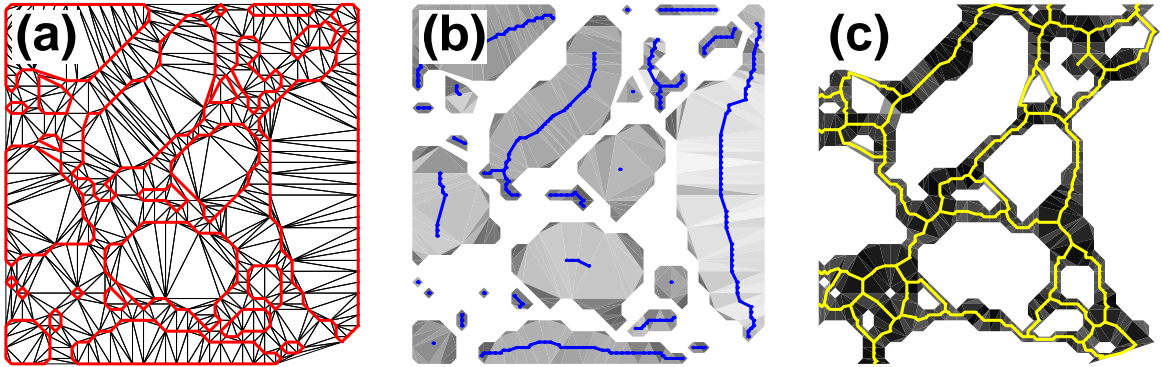,width=16.2cm} 
   \end{tabular}
   \vspace*{-0.5cm}
   \end{center}
   \caption[example] { 	  
   (a) Delaunay tesselation,
   (b) grains and pruned CAT skeleton,
   (c) binder and pruned CAT skeleton, (see text). 
   } 
   \end{figure} 
 
(among 256 grey levels) for cutting a torso, thus leading to grain 
separation. In
Fig. 6.c we show the resulting refined grain contours. We note, that it is 
neccessary for the shape manipulation to use the unpruned CAT skeleton.

\subsection{Material Statistics} 

After the grain separation it is in general neccessary to generate a new CDT
of the refined contours (cf. Fig. 7.a). In the subsequent CAT our computer
code \cite{Geofilt1} generates a connectivity hierarchy of the triangular
chain complexes, which allows the efficient pruning of morphologically
insignificant shape features. The results are the pruned skeleton for
grains (cf. Fig. 7.b) and the pruned skeleton for the polymeric binder
(cf. Fig. 7.c). We note, that the CAT skeleton of a grain can be sometimes
pruned to a single point. After pruning, each connected arc of the attributed 
CAT skeletons can be regarded as the axis of a single particle (grain or
binder particle). This allows for the computation of area, length, width,
location, orientation, etc. for each particle. In case of present holes
in a single HMX grain we even are able to remove them through the knowledge
of the morphology of a single grains CAT skeleton. Of course, this requires
a third application of CDT and CAT, respectively.

Figs. 8. and 9. show the processed images for Figs. 3 and 4. E.g., we find
the following grain numbers when going from Fig. 8.a over Fig. 8.b and 
Fig. 9.a to Fig. 9.b: 912, 919, 624, and 534.
Correspondingly, the two-dimensional HMX crystal percentages are:
75\%, 72\%, 74\%, and 71\%. Obviously, although the ratio between grains
and polymer binder are more or less insensitive to the choice of image
segmentation, there are apparently significant differences in 

   \begin{figure}[b]
   \begin{center}
   \vspace*{-0.5cm}
   \begin{tabular}{c}
   \psfig{figure=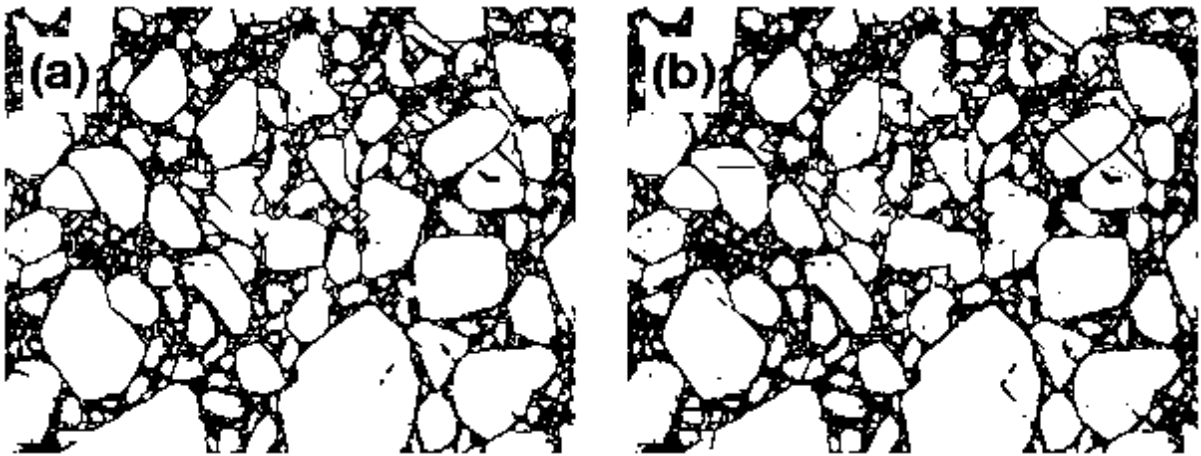,width=16.2cm} 
   \end{tabular}
   \vspace*{-0.5cm}
   \end{center}
   \caption[example] { 	  
   processed original images, which are initially segmented by using
   (a) spectral range,
   (b) PCNN, (see text). 
   } 
   \end{figure} 
 
   \begin{figure}[t]
   \begin{center}
   \vspace*{-0.5cm}
   \begin{tabular}{c}
   \psfig{figure=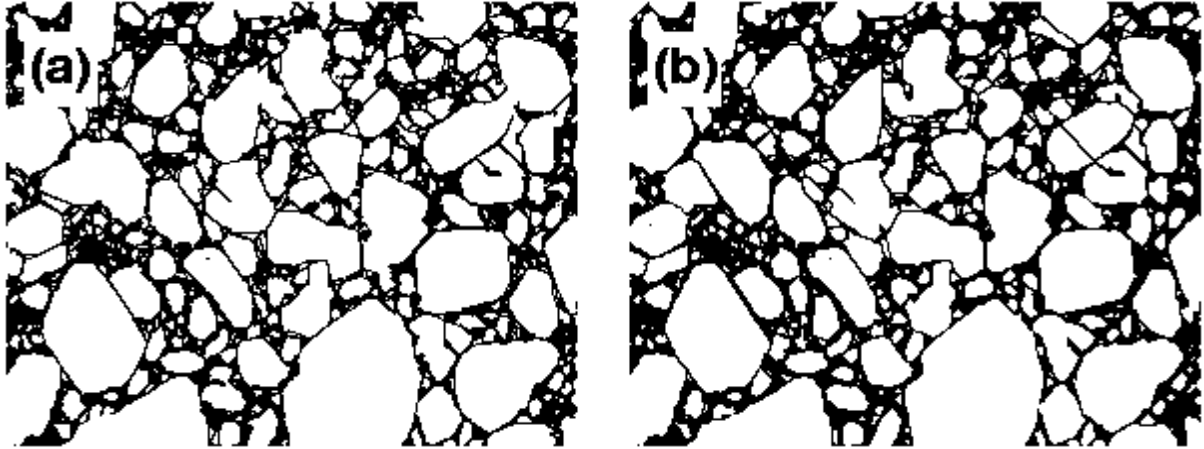,width=16.2cm} 
   \end{tabular}
   \vspace*{-0.5cm}
   \end{center}
   \caption[example] { 	  
   processed smoothed images, which are initially segmented by using
   (a) spectral range,
   (b) PCNN, (see text). 
   } 
   \end{figure} 
 
the grain 
numbers. Further experimental analysis should help to determine, which
image segmentation is the most favorable for the computer vision micrograph
analysis of explosive materials such as the PBX 9501 considered here.

\section{SUMMARY}

In summary, we have presented a method of image decomposition for
micrographs into meaningful parts such as single HMX and binder grains. 
We have demonstrated that our algorithms are capable of grain separation 
and the efficient computation of statistical features of granular materials.

\acknowledgments     
 
This work has been supported by the Department of Energy.




\begin{thebibliography}{99}   

\bibitem{Matheron65} 
G. Matheron, ``Les variables r\'egionalis\'ees et leur 
estimation'', Doctorate thesis, {\em Appl. Sci.}, Masson, Paris, 1965.

\bibitem{Serra65} 
J. Serra, ``L'analyse des textures par la g\'eom\'etrie 
al\'eatoire'', {\em Compte-rendu du Comit\'e Scientifique de 
l'IRSID}, 1965.

\bibitem{SkourikhineThis}
A.~N. Skourikhine, L. Prasad, B.~R. Schlei, ``A neural network for
image segmentation'',  in {\em Mathematical Imaging}, {\em Proc. of 
SPIE's 45$^{th}$ Annual International Symposium, San Diego, CA, SPIE} 
Vol. {\bf 4120}, 2000.

\bibitem{Prasad97}
L. Prasad, ``Morphological Analysis of Shapes'', {\em CNLS Newsletter}, No. 
{\bf 139}, LALP-97-010-139, Center for Nonlinear Studies, Los Alamos 
National Laboratory, July `97.

\bibitem{PrasadThis}
L. Prasad, R. Rao, ``A Geometric Transform for Shape Feature Extraction'', 
in {\em Mathematical Imaging}, {\em Proc. of SPIE's 45$^{th}$ Annual 
International Symposium, San Diego, CA, SPIE} Vol. {\bf 4117}, 2000. 
 
\bibitem{Skidmore97}
C.~B. Skidmore, D.~S. Phillips, N.~B. Crane, ``Microscopical Examination 
of Plastic-Bonded Explosives'', {\em Microscope}, {\bf 45}(4), 
pp.~127--136, 1997.

\bibitem{Diconex1}
B.~R. Schlei, "DICONEX -- Dilated Contour Extraction Code, Version 1.0", 
{\em Los Alamos Computer Code} LA-CC-00-30, Los Alamos National Laboratory, 
2000; for more detail check the website 
{\tt http://www.nis.lanl.gov/$\sim$bschlei/eprint.html}.

\bibitem{Schlei00}
B.~R. Schlei, L. Prasad, "A Parallel Algorithm for Dilated Contour 
Extraction from Bilevel Images", {\em Los Alamos Preprint} LA-UR-00-309, 
Los Alamos National Laboratory, cs.CV/0001024, 2000.

\bibitem{Geofilt1}
B.~R. Schlei, "GEOFILT -- Geometric Filtering Code, Version 1.0", 
{\em Los Alamos Computer Code} LA-CC-00-31, Los Alamos National Laboratory, 
2000; for more detail check the website 
{\tt http://www.nis.lanl.gov/$\sim$bschlei/eprint.html}.

\end{thebibliography}
  \end{document}